\pdfoutput=1

\documentclass[11pt]{article}

\usepackage[]{ACL2023}

\usepackage{times}
\usepackage{latexsym}

\usepackage{microtype}
\usepackage{enumerate}
\usepackage{makecell}
\usepackage{booktabs}
\usepackage{multirow}
\usepackage{xcolor}
\usepackage{subcaption}

\usepackage[T1]{fontenc}

\usepackage[utf8]{inputenc}
\newcommand\blfootnote[1]{%
  \begingroup
  \renewcommand\thefootnote{}\footnote{#1}%
  \addtocounter{footnote}{-1}%
  \endgroup
}
\usepackage{microtype}

\usepackage{inconsolata}
\usepackage{graphicx}

\title{A Text-to-Text Model for Multilingual Offensive Language Identification}

\author{Tharindu Ranasinghe\textsuperscript{*} \\
  Aston University \\
  Birmingham, UK\\
  \texttt{t.ranasinghe@aston.ac.uk} \\\And
  Marcos Zampieri\textsuperscript{*} \\
  George Mason University \\
  Fairfax, VA, USA \\
  \texttt{mzampier@gmu.edu} \\}

\begin{document}
\maketitle
\begin{abstract}
The ubiquity of offensive content on social media is a growing cause for concern among companies and government organizations. Recently, transformer-based models such as BERT, XLNET, and XLM-R have achieved state-of-the-art performance in detecting various forms of offensive content (e.g. hate speech, cyberbullying, and cyberaggression). However, the majority of these models are limited in their capabilities due to their encoder-only architecture, which restricts the number and types of labels in downstream tasks. Addressing these limitations, this study presents the first pre-trained model with encoder-decoder architecture for offensive language identification with text-to-text transformers (T5) trained on two large offensive language identification datasets; SOLID and CCTK. We investigate the effectiveness of combining two datasets and selecting an optimal threshold in semi-supervised instances in SOLID in the T5 retraining step. Our pre-trained T5 model outperforms other transformer-based models fine-tuned for offensive language detection, such as fBERT and HateBERT, in multiple English benchmarks. Following a similar approach, we also train the first multilingual pre-trained model for offensive language identification using mT5 and evaluate its performance on a set of six different languages (German, Hindi, Korean, Marathi, Sinhala, and Spanish). The results demonstrate that this multilingual model achieves a new state-of-the-art on all the above datasets, showing its usefulness in multilingual scenarios. Our proposed T5-based models will be made freely available to the community. 
\blfootnote{\bf WARNING: This paper contains examples that are offensive in nature.}
\end{abstract}

\section{Introduction}

The widespread of offensive posts on social media platforms can have detrimental effects on users' mental health among other undesirable consequences. The relation between offensive language and mental health along with potential risks of self harm and depression has been widely addressed by previous studies \cite{bonanno2013cyber,bannink2014cyber,bucur-etal-2021-exploratory}. To address this important issue, one of the most commonly employed strategies is to train systems to identify offensive content \cite{pavlopoulos-etal-2021-semeval} mitigating its spread on social media platforms. By proactively identifying potentially harmful content, social media platforms aim to establish a safer and more inclusive environment for all users.
\blfootnote{*The two authors contributed equally to this work.}

Early approaches to identifying offensive language ranged from classical machine learning models, such as support vector machines \cite{malmasi-zampieri-2017-detecting,malmasi2018challenges}, to deep learning models based on word embeddings \cite{hettiarachchi-ranasinghe-2019-emoji}. With the introduction of BERT \cite{devlin2019bert}, transformer models have shown excellent results in offensive language identification \cite{zia2022improving}. More recently, domain-specific language models for offensive language identification, such as fBERT \cite{sarkar2021fbert}, HateBERT \cite{caselli2021hatebert}, and ToxicBERT\footnote{ToxicBERT is available at \url{https://huggingface.co/unitary/toxic-bert}}. have provided state-of-the-art in multiple offensive language identification benchmarks.

The aforementioned models can be grouped into two main categories following their training strategies. Models such as ToxicBERT have been trained using a classification objective by adding a classification layer on top of a BERT model and training on a large offensive language dataset. A clear limitation of this approach is that the trained model can only predict the classes that appear on the dataset. On the other hand, models such as fBERT \cite{sarkar2021fbert} and HateBERT \cite{caselli2021hatebert} have been trained with a masked language modelling (MLM) objective. fBERT \cite{sarkar2021fbert} has been trained on the offensive tweets in the SOLID \cite{rosenthal2020largescale} dataset, while HateBERT \cite{caselli2021hatebert} has been trained on banned posts from Reddit Abusive Language dataset. The MLM strategy is not dependent on the number of classes present in the dataset. However, it is not possible to concatenate two datasets annotated with different annotation taxonomies under this strategy without mapping them into a common label (e.g. a general offensive class). This is a critical issue in offensive language identification as different datasets use different annotation schemes and problem formulations (e.g. hate speech, offensive, toxic, profanity). As a result, MLM based models are only trained on one dataset, which can limit their capabilities. 

To address this important shortcoming, we introduce FT5, a pre-trained T5 model \cite{raffel2020exploring} trained on two large-scale offensive language identification datasets. Since T5 follows a text-to-text approach, it does not rely on a classification layer. Therefore T5 \cite{raffel2020exploring} can be used to train an offensive language identification model using different datasets without relying on the number of classes. We show that the proposed FT5 outperforms the plain T5 implementation as well as HateBERT \cite{caselli2021hatebert} and fBERT \cite{sarkar2021fbert} on various offensive and hate speech detection tasks. To the best of our knowledge, this is the first pre-trained offensive language identification model based on T5. 

All the previous models, such as ToxicBERT, HateBERT \cite{caselli2021hatebert} and fBERT \cite{sarkar2021fbert} only supports English and training a large language model using similar approaches in low-resource languages can be difficult due to data scarcity. In this paper, we address this limitation by training a multilingual offensive language model, mFT5, which uses mT5 \cite{xue-etal-2021-mt5} as the base model. the results confirm that fine-tuned mFT5 produces state-of-the-art results in six languages, outperforming strong transformer-based models. To the best of our knowledge, mFT5 is the first multilingual model on offensive language opening exciting avenues for a multitude of languages.

The contributions of this paper are as follows:

\begin{enumerate}
    \item An empirical evaluation of semi-supervised learning techniques that can be applied to train text-to-text models such as T5 \cite{raffel2020exploring} and mT5 \cite{xue-etal-2021-mt5} in offensive language identification

    \item A comprehensive evaluation of the effect of combining different datasets in pre-training text-to-text models.

    \item The first-ever cross-lingual evaluation of mT5 \cite{xue-etal-2021-mt5} model in both high-resource and low-resource language settings. 

    \item The release of the FT5 and mFT5 made freely available to the research community, which are high-performing, state-of-the-art pre-trained models based on T5 for English and multilingual offensive language identification\footnote{\url{https://github.com/TharinduDR/FT5}}. 

\end{enumerate}

\section{Related Work}

\subsection{Offensive Language Identification} 

The use of large pre-trained transformer models has become prevalent in NLP. This includes the development of various offensive language identification systems, which are based on transformer architectures, such as BERT \cite{devlin2019bert}. These systems have demonstrated top performance in well-known competitions such as HASOC \cite{hasoc2019}, HatEval \cite{basile-etal-2019-semeval}, OffensEval \cite{offenseval,zampieri-etal-2020-semeval}, TRAC \cite{kumar-etal-2018-benchmarking}, and TSD \cite{pavlopoulos2021semeval}. Many of these competitions feature multilingual datasets opening opportunities for the use of cross-lingual models \cite{ranasinghe-etal-2020-multilingual,ranasinghe2021multilingual,nozza2021exposing}. The outstanding results achieved by these systems provide concrete evidence that pre-trained transformer models are well-suited for detecting offensive content in both monolingual and multilingual settings.

User-generated content and offensive language online possess unique characteristics that are often not adequately captured by models trained on standard texts. Consequently, research has focused on the task of fine-tuning pre-trained models specifically for this challenging domain. There are several transformer models such as HateBERT \cite{caselli2021hatebert}, fBERT \cite{sarkar2021fbert} built for this purpose. In this study, we address the aforementioned limitations of fine-tuned transformer-based models. We propose the first multilingual domain-specific pre-trained offensive language identification model. 

\subsection{T5 Models}

T5 models introduced by \citet{raffel2020exploring} have been widely used in many NLP tasks such as text classification \cite{Bird2023}, semantic similarity \cite{ni-etal-2022-sentence} and named entity recognition \cite{tavan-najafi-2022-marsan}. As the T5 architecture follows a text-to-text approach, multi task learning can be used to improve the results with t5 \cite{raffel2020exploring}. Following the initial T5 model, multilingual T5 (mT5) models have also been proposed by \citet{xue-etal-2021-mt5} which has provided excellent results in multilingual benchmarks. Several studies have used T5 in offensive language identification \cite{sabry2022hat5, adewumi2022t5}. However, these studies only fine-tune the general T5 models for offensive language identification. To the best of our knowledge, this paper presents the first pre-trained domain specific T5 model for offensive language identification. 

\section{Methodology}
\label{sec:methods}

 \paragraph{Training Data} In this study, we use two large offensive language identification datasets to retrain the T5 models; SOLID \cite{rosenthal2020largescale} with over $9$ million English tweets and CCTK with over $1.8$ million posts from the civil comments platform. SOLID was the official dataset of SemEval-2020 Task 12 (OffensEval) \cite{zampieri-etal-2020-semeval}. SOLID's annotation follows the OLID taxonomy \cite{zampieri-etal-2019-predicting}, which uses its level A (offensive vs non-offensive). Each instance in SOLID has been annotated semi-automatically with the mean and the standard deviation (STD) of the value for offensiveness predicted by four different machine learning models. CCTK was released for the Jigsaw Unintended Bias in Toxicity Classification Kaggle competition\footnote{\url{https://www.kaggle.com/c/jigsaw-unintended-bias-in-toxicity-classification}}. Each instance in CCTK has been annotated with one of the binary labels (toxic vs not toxic). Finally, we used multiple datasets for testing presented in Sections \ref{sec:english} and \ref{sec:multi}.

\begin{figure}
\centering
\includegraphics[scale=0.53]{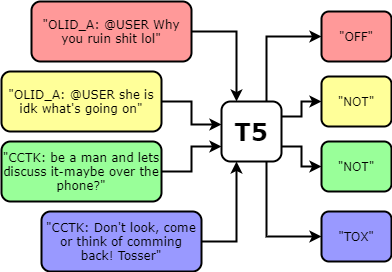}
\caption{T5/ MT5 pre-training process}
\label{fig:fbert}
\end{figure}

 \paragraph{Retraining T5} We select the \textit{t5-large}\footnote{t5-large model is available in HuggingFace at \url{https://huggingface.co/t5-large}} \cite{raffel2020exploring} and train it using the instances from SOLID \cite{rosenthal2020largescale} and CCTK. For the instances in SOLID, the input texts to the model were tweets, and the output texts were "OFF" if the mean value in SOLID is above 0.5 and "NOT" otherwise. We used different thresholds (0.05, 0.1, 0.15 and 0.2) for the STD value to filter the most confident examples from SOLID. For each threshold, we consider appending/ not appending the CCTK dataset. For the instances in CCTK, the input texts to the model were the posts, and the output texts were "TOX" if the text is toxic and "NOT" if they are not toxic. As shown in Figure \ref{fig:fbert} we use "OLID\_A" prefix for SOLID instances and "CCTK" prefix for CCTK instances. To create mFT5, we select \textit{mt5-large} \cite{xue-etal-2021-mt5}\footnote{mt5-large model is available in HuggingFace at \url{https://huggingface.co/google/mt5-large}} and repeat the same process. For both models, we use the same configurations; a batch-size of 16, Adam optimiser with learning rate $1\mathrm{e}{-4}$, and a linear learning rate warm-up over 10\% of the training data and trained the models over ten epochs. We use a cluster of four GeForce RTX 3090 GPUs to train the models. 

\section{English Experiments and Results}
\label{sec:english}

\begin{figure}[!ht]
\centering
\includegraphics[scale=0.4]{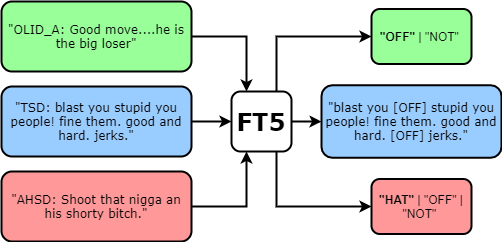}
\caption{FT5 fine-tuning process for different tasks.}
\label{fig:ft5_fine}
\end{figure}

To determine the effectiveness and portability of the trained FT5, we conducted a series of experiments using benchmark datasets in English and compared our model with a general-purpose T5 model. We used the same set of configurations for all the datasets evaluated in order to ensure consistency between all the experiments. This also provides a good starting configuration for researchers who intend to use FT5 on a new dataset. For the sentence-level tasks; the input to the model is the text and the output is the related label. For the token level tasks the input to the model is the text and the output is the text with "[OFF]" placeholders infront of the offensive tokens as shown in Figure \ref{fig:ft5_fine}. 

For each dataset, we used different task specific prefixes (OLID\_A for OLID, TSD for toxic spans detection etc.). We used a batch-size of eight, Adam optimizer with learning rate $1\mathrm{e}{-4}$, and a linear learning rate warm-up over 10\% of the training data. During the training process, the parameters of the transformer model were updated. The models were trained using only training data. Furthermore, they were evaluated while training using an evaluation set that had one fifth of the rows in training data. We performed early stopping if the evaluation loss did not improve over ten evaluation steps. All the models were trained for three epochs. These experiments were also conducted in a GeForce RTX 3090 GPU. All the experiments were conducted for ten times and we report the mean and standard deviation for each experiment.

\paragraph{AHSD} In fine-grain aggression detection, classifying offensive language and hate speech is challenging. Hate speech contains explicit instances targeted towards a specific group of people intended to degrade or insult. \citet{davidson2017} compiled a $24,783$ English tweets dataset annotated with one of three labels -- ``hate speech'', ``only offensive'', and ``neither''. The dataset contains $1,430$ hate speech, $19,190$ only offensive, and $4,163$ instances that are neither. We further split the dataset into training and test sets in a 4:1 ratio.  

\paragraph{OLID} We use OLID, the official dataset for OffensEval 2019 \cite{offenseval}, one of the the most popular offensive language identification shared tasks. The dataset has $13,240$ training and $860$ test instances. There are $4,400$ and $240$ offensive posts in the training and test dataset, respectively. For the experiment, we chose sub-task A, a binary classification task between offensive and non-offensive posts.

\paragraph{TRAC} TRAC was released for TRAC shared task 2020 \cite{kumar-etal-2020-evaluating}. The dataset has $4200$ training and $1200$ test instances with three classes: overtly aggressive, covertly aggressive and non-aggressive. TRAC is the most heterogeneous dataset we used in terms of data sources containing posts from Facebook, Twitter, and YouTube. 

\paragraph{TSD} For token-level prediction we use TSD, released within the scope of SemEval-2021 Task 5: Toxic Spans Detection for English \cite{pavlopoulos-etal-2021-semeval}. The dataset contains 10,000 posts (comments) from the publicly available Civil Comments dataset \cite{borkan2019nuanced}. If a post is toxic, it has been annotated for its toxic spans. 

\paragraph{HateX} HateXplain dataset \cite{mathew2020hatexplain} was also for offensive language identification at the token level. The dataset contains $11535$ training and $3844$ testing instances from GAB and twitter. We only used the word level annotations. 

\renewcommand{\arraystretch}{1.2}
\begin{table*}[t]
\begin{center}
\scalebox{0.85}{
\begin{tabular}{l c l | c c c | c c} 
\toprule
&  &  & \multicolumn{3}{c}{\bf Sentence-level} & \multicolumn{2}{c}{\bf Token-level} \\
\cmidrule(r){4-6}\cmidrule(r){7-8}
{\bf Train Dataset(s) } &  \bf STD & \bf Inst. & \bf AHSD & \bf OLID & \bf TRAC & \bf TSD & \bf HateX \\
\midrule
\multirow{ 4}{*}{SOLID} &  0.05 & 18,169 & 0.832 \textpm 0.009  & 0.807\textpm0.006 & 0.849 \textpm 0.005 & 0.542 \textpm 0.004 & 0.801 \textpm 0.006 \\
&  0.1  & 215,602 & 0.846 \textpm 0.005 & 0.819\textpm0.002 & 0.854 \textpm 0.003 & 0.601 \textpm 0.005 & 0.816 \textpm 0.005 \\
&  0.15  & 1,282,474 & 0.870 \textpm 0.005 & 0.813\textpm0.003 & \textbf{0.870 \textpm 0.002} & 0.591 \textpm 0.008 & 0.801 \textpm 0.006 \\
&  0.2  & 6,595,397 & 0.859 \textpm 0.003 & 0.805\textpm0.005 &  0.865 \textpm 0.005 & 0.561 \textpm 0.005 & 0.795 \textpm 0.006 \\
\midrule
\multirow{ 4}{*}{SOLID+CCTK} &  0.05 & 1,823,043 & 0.865 \textpm 0.004  & 0.819\textpm0.003 & 0.859 \textpm 0.007 & 0.609 \textpm 0.004 & 0.815 \textpm 0.004 \\
&  0.1  & 2,020,476 & \textbf{0.886 \textpm 0.004} &  \textbf{0.823 \textpm 0.002}  & 0.869 \textpm 0.005 & \textbf{0.648 \textpm 0.005} & \textbf{0.825 \textpm 0.004}\\
&  0.15  & 3,087,348 & 0.872 \textpm 0.005 & 0.816\textpm0.004 & 0.864 \textpm 0.005 & 0.605 \textpm 0.006 & 0.812 \textpm 0.007 \\
&  0.2  & 8,400,271 & 0.868 \textpm 0.005 & 0.809\textpm0.006 & 0.858 \textpm 0.006 & 0.589 \textpm 0.012 & 0.808 \textpm 0.009 \\
\midrule
CCTK &  NA & 1,804,874 & 0.832 \textpm 0.009  & 0.813\textpm0.006 & 0.842 \textpm 0.005 & 0.595\textpm 0.004 & 0.809 \textpm 0.006 \\
\bottomrule
\end{tabular}
}
\end{center}
\caption{FT5 results for different sentence-level and token-level offensive language detection benchmarks. } 
\label{tab:ft5_results}
\end{table*}

\subsection{Sentence-level Offensive Language Identification}
To evaluate the sentence-level tasks, we used the macro F1 score computed on predicted sentence labels and gold sentence labels. 

We present the results for the SOLID data selection thresholds and data augmentation with CCTK in Table \ref{tab:ft5_results} in terms of F$_1$ Macro. For most of the datasets tested, the 0.1 STD SOLID threshold combined with the CCTK provided the best results. Having a large number of training instances provides better results up to a certain STD threshold in SOLID, and results do not improve with adding further training instances. Furthermore, the results show that the combination of CCTK and SOLID provides better results than having one dataset in the training set. This confirms our previous assumption that T5 can take advantage of multiple datasets via text to text transfer learning. 

We select the T5 model retrained on SOLID filtered with 0.1 STD combined with the CCTK dataset as the FT5 model, which provided the best result in most of the datasets. We then compare the performance of FT5 with fBERT and HateBERT.

\begin{table}[!ht]
\centering
\scalebox{.9}{
\begin{tabular}{llc}
\hline
\bf Dataset & \textbf{Model} & \textbf{Macro F1} \\ \hline
 \multirow{ 4}{*}{AHSD} &  \bf FT5 & \bf 0.886\textpm0.004 \\ 
&  fBERT & 0.878\textpm0.005 \\ 
& HateBERT & 0.846\textpm0.009 \\
& T5 & 0.821\textpm0.012 \\
\hline
 \multirow{ 4}{*}{OLID} &  \bf FT5 & \bf 0.823\textpm0.002 \\ 
&  fBERT & 0.810\textpm0.005 \\ 
& HateBERT & 0.803\textpm0.009 \\
& T5 & 0.775\textpm0.006 \\
\hline
 \multirow{ 4}{*}{TRAC} &  \bf FT5 & \bf 0.869\textpm0.003 \\ 
&  fBERT & 0.859\textpm0.005 \\ 
& HateBERT & 0.848\textpm0.006 \\
& T5 & 0.846\textpm0.010 \\
\bottomrule
\end{tabular}
}
\caption{The test set macro F$_1$ scores for sentence-level datasets and models. Results are ordered by performance. Best results are shown in bold font.}
\label{table:ft5_sentence}
\end{table}

\noindent As can be seen in Table \ref{table:ft5_sentence}, FT5 outperforms fBERT and HateBERT in all of the datasets. Since the datasets contain offensive language identification, fine grained offensive language identification and fine-grained aggression identification, we can validate the effectiveness of the proposed FT5 model for offensive and aggressive language sentence-level classification tasks. 

\subsection{Token-level Offensive Language Identification}

Multiple studies on token-level offensive language identification has discussed the need for accurate token-level predictions for improved model explainability \cite{mathew2020hatexplain,zampieri-etal-2023-target}. Motivated by recent studies, we investigate our model performance on token-level offensive language identification. The token-level tasks were evaluated using the macro F1 score computed on predicted character offsets and gold character offsets \cite{da-san-martino-etal-2019-fine}.

\begin{table}[!ht]
\centering
\scalebox{.9}{
\begin{tabular}{llc}
\hline
\bf Dataset & \textbf{Model} & \textbf{Macro F1} \\ \hline
 \multirow{ 4}{*}{TSD} &  \bf FT5 & \bf0.648\textpm0.012 \\ 
&  fBERT & 0.530\textpm0.021 \\
& T5 & 0.421\textpm0.019 \\
& HateBERT & 0.410\textpm0.027 \\

\hline
 \multirow{ 4}{*}{HateX} &  \bf FT5 & \bf 0.825\textpm0.008 \\ 
&  fBERT & 0.812\textpm0.009 \\ 
& HateBERT & 0.792\textpm0.016 \\
& T5 & 0.775\textpm0.025 \\
\bottomrule

\end{tabular}
}
\caption{The test set macro F$_1$ scores for sentence-level datasets and models. Results are ordered by performance. Best results are shown in bold font.}
\label{table:ft5_word}
\end{table}

\noindent FT5 outperforms fBERT and HateBERT in all of the token-level offensive language identification datasets too as can be seen in Table \ref{table:ft5_word}. There is a clear improvement with the TSD dataset where the FT5 model outperforms the fBERT model by 0.11 macro F1 score which is over a $20\%$ boost.

\section{Multilingual Experiments and Results}
\label{sec:multi}
To determine the effectiveness and portability of our multilingual model; mFT5, we conducted a series of experiments using benchmark datasets covering high-resource, mid-resource and low-resource languages. These datasets are summarised in Table \ref{tab:data}. We used the same set of configurations we used for English experiments. The models were trained using the training set and evaluated on the test sets of each dataset. 

\begin{table*}[t]
\begin{center}
\scalebox{0.75}{
\begin{tabular}{ c|l|c|c|c|c } 
 \hline
 \textbf{Language} & \textbf{Source(s)}  &  \textbf{Train, dev, test size} & \textbf{Labels} & \textbf{Sentence-level} & \textbf{Token-level} \\ 
  \hline
   German \cite{risch-etal-2021-overview} & Facebook &   2076, 519, 649 &  \makecell[l]{ Toxic \\ Not-toxic } & TOX, NOT &  NA\\
   \hline
     Spanish \cite{plaza-del-arco-etal-2021-offendes} & \makecell[l]{Twitter \\ Instagram \\ Youtube}  &   30163, 7540, 9425 &  \makecell[l]{ Offensive individual target \\  Offensive group target \\ Offensive other target \\ Expletive language \\ Non-offensive } & OFF, NOT &  NA\\
      \hline
     Hindi \cite{hasoc2019} & Twitter &  5120, 1280, 1600 &  \makecell[l]{ Offensive \\ Not offensive } & OFF, NOT &  NA\\
     \hline
      Korean \cite{jeong2022kold} & \makecell[l]{Naver  \\ YouTube} &   25876, 6468, 8085 &  \makecell[l]{ Offensive \\ Not offensive } & OFF, NOT &  Available\\
      \hline
      Sinhala \cite{ranasinghe2022sold} & Twitter &   6000, 1500, 2500 &  \makecell[l]{ Offensive \\ Not offensive } & OFF, NOT &  Available\\
       \hline
      Marathi \cite{gaikwad-etal-2021-cross} & Twitter &   2889, 722, 510 &  \makecell[l]{ Offensive \\ Not offensive } & OFF, NOT &  NA\\
   \hline
\end{tabular}
}
\end{center}
\caption{Datasets that were used to evaluate mFt5 model. \textbf{Source} column displays the platform data extracted,\textbf{Train, dev, test size} column shows the number of instances of the train, dev and test sets. \textbf{Label} column shows the original labels and \textbf{Sentence-level} column show the output label in sentence-level experimenst discussed in Section \ref{sec:multi}. \textbf{Token-level} column shows the availability of the token-level data. } 
\label{tab:data}
\end{table*}

\renewcommand{\arraystretch}{1.2}
\begin{table*}[t]
\begin{center}
\scalebox{0.8}{
\begin{tabular}{l c l | c c | c c | c c} 
\toprule
&  &  & \multicolumn{2}{c}{\bf High Resource} & \multicolumn{2}{c}{\bf Mid Resource} & \multicolumn{2}{c}{\bf Low Resource} \\
\cmidrule(r){4-5}\cmidrule(r){6-7}\cmidrule(r){8-9}
{\bf \makecell{Train \\ Dataset(s) }} &  \bf STD & \bf Inst. & \bf German & \bf Spanish & \bf Hindi & \bf Korean & \bf Sinhala & \bf Marathi \\
\midrule
\multirow{ 4}{*}{SOLID} &  0.05 & 18,169 & 0.567 \textpm 0.010  & 0.832 \textpm 0.009 & 0.799 \textpm 0.005 & 0.735 \textpm 0.004 & 0.748 \textpm 0.008 & 0.789 \textpm 0.016 \\
&  0.1  & 215,602 & 0.601 \textpm 0.002 & 0.846 \textpm 0.005 & 0.807 \textpm 0.003 & 0.776 \textpm 0.005 & 0.756 \textpm 0.008 & 0.825 \textpm 0.009 \\
&  0.15  & 1,282,474 & 0.598 \textpm 0.007 & 0.870 \textpm 0.005 & 0.839 \textpm 0.002 & 0.781 \textpm 0.008 & 0.784 \textpm 0.007 & \textbf{0.858 \textpm 0.006} \\
&  0.2  & 6,595,397 & 0.588 \textpm 0.006 & 0.859 \textpm 0.003 &  0.825 \textpm 0.005 & 0.757 \textpm 0.005 & 0.766 \textpm 0.008 & 0.844 \textpm 0.010 \\
\midrule
\multirow{ 4}{*}{\makecell{SOLID \\ +CCTK}} &  0.05 & 1,823,043 & 0.611 \textpm 0.004  & 0.852\textpm0.005 & 0.859 \textpm 0.007 & 0.769 \textpm 0.004 & 0.812\textpm0.016 & 0.835 \textpm 0.009 \\
&  0.1  & 2,020,476 & \textbf{0.653\textpm0.031} &  \textbf{0.886 \textpm 0.004} & \textbf{0.845 \textpm 0.005}& \textbf{0.799\textpm0.004} & \textbf{0.856\textpm0.007} & 0.854\textpm0.006 \\
&  0.15  & 3,087,348 & 0.642\textpm 0.005 & 0.872 \textpm 0.005 & 0.822 \textpm 0.005 & 0.778 \textpm 0.006 & 0.856 \textpm 0.006 & 0.849\textpm 0.008 \\
&  0.2  & 8,400,271 & 0.611 \textpm 0.005 & 0.843\textpm0.012 & 0.818 \textpm 0.006 & 0.765 \textpm 0.012 & 0.836 \textpm 0.006 & 0.811\textpm 0.005 \\
\midrule
CCTK &  NA & 1,804,874 & 0.628 \textpm 0.009  & 0.829\textpm0.006 & 0.825 \textpm 0.005 & 0.775\textpm 0.004 & 0.796 \textpm 0.005 & 0.801 \textpm 0.003 \\
\bottomrule
\end{tabular}
}
\end{center}
\caption{mFT5 results for different multilingual offensive language detection benchmarks. } 
\label{tab:mft5_results}
\end{table*}

\subsection{Sentence-level Offensive Language Identification}

For sentence-level offensive language identification, we mapped the labels of each dataset to its closet annotation scheme as we did for English benchmarks; OLID level A (offensive, not offensive) or CCTK (toxic, non toxic). Following this, Spanish, Hindi, Korean, Sinhala and Marathi labels were mapped to OLID level A, and German labels were mapped to CCTK as shown in \textit{sentence-level} column in Table \ref{tab:data}. We use "OLID\_A" prefix for Spanish, Hindi, Korean, Sinhala and Marathi instances and "CCTK" prefix for German instances. In the training process, we started with different pre-trained models on the configurations described in Section \ref{sec:methods}. The input to the model is the text preceded by the relevant prefix, and the output is the related label. We performed individual experiments for each language separately.

We present the results for the SOLID data selection thresholds and data augmentation with CCTK in Table \ref{tab:mft5_results} in terms of F$_1$ Macro for each test set. For most of the datasets tested, the 0.1 STD SOLID threshold combined with the CCTK provided the best results. Having a large number of training instances provides better results up to a certain STD threshold in SOLID, and results do not improve with adding further training instances. Furthermore, the results show that the combination of CCTK and SOLID provides better results than having one dataset in the training set. This confirms our previous assumption that T5 can take advantage of multiple datasets via text to text transfer learning. We select the T5 model retrained on SOLID filtered with 0.1 STD combined with the CCTK dataset as the FT5 model, which provided the best result in five out of six datasets.

We then compare the performance of mFT5 with mBERT and XLM-R base models. These models are trained on the training set of each dataset by adding a classification layer on top of the transformer model. As shown in Table  \ref{tab:fmt5_coarse}, mFT5 outperforms mBERT and XLM-R in all of the datasets. Since these datasets contain high-resource and low-resource languages as well as data from different social media platforms, we can validate the effectiveness of the proposed mFT5 model for offensive language identification in multiple languages and platforms. 

\begin{table}[!ht]
\centering
\scalebox{.9}{
\begin{tabular}{llc}
\hline
\bf Dataset & \textbf{Model} & \textbf{Macro F1} \\ \hline
 \multirow{ 4}{*}{German} &  \bf mFT5 & \bf 0.653\textpm0.031  \\ 
&  XLM-R & 0.621\textpm0.023 \\ 
& mBERT & 0.572\textpm0.013\\
&  mFT5* & 0.438\textpm0.015 \\ 
& mT5 & 0.398\textpm0.016 \\
\hline
 \multirow{ 4}{*}{Spanish} &  \bf mFT5 & \bf 0.886\textpm0.004 \\ 
&  XLM-R &  0.853\textpm0.005  \\ 
& mBERT &  0.821\textpm0.008   \\
&  mFT5* & 0.785\textpm0.005 \\ 
& mT5 & 0.626\textpm0.027 \\
\hline
 \multirow{ 4}{*}{Hindi} &  \bf mFT5 & \bf 0.845\textpm0.003 \\ 
&  XLM-R & 0.811\textpm0.007 \\ 
& mBERT & 0.798\textpm0.006 \\
&  mFT5* & 0.745\textpm0.007 \\ 
& mT5 &  0.612\textpm0.012 \\
\hline
 \multirow{ 4}{*}{Korean} &  \bf mFT5 & \bf 0.799\textpm0.004 \\ 
&  XLM-R & 0.765\textpm0.006 \\ 
& mBERT &  0.755\textpm0.008 \\
&  mFT5* & 0.736\textpm0.008 \\ 
& mT5 & 0.715\textpm0.005 \\
\hline
 \multirow{ 4}{*}{Sinhala} &  \bf mFT5 & \bf 0.856\textpm0.007 \\ 
&  XLM-R &  0.834\textpm0.005  \\
&  mFT5* & 0.746\textpm0.009 \\ 
& mT5 & 0.538\textpm0.029 \\
& mBERT & 0.531\textpm0.013 \\

\hline
 \multirow{ 4}{*}{Marathi} &  \bf mFT5 & \bf 0.854\textpm0.006 \\ 
 &  XLM-R & 0.843\textpm0.003 \\ 
& mBERT & 0.821\textpm0.006  \\
&  mFT5* & 0.708\textpm0.012 \\ 
& mT5 & 0.421\textpm0.017 \\
\bottomrule
\end{tabular}
}
\caption{The test set macro F$_1$ scores for coarse-grained offensive language detection. Results are ordered by performance. The best results are shown in bold font. }
\label{tab:fmt5_coarse}
\end{table}

\noindent \textbf{Zero-shot Offensive Language Identification} - We also experimented with zero-shot cross-lingual offensive language identification with the mFT5 model. With this setting, we did not train the mFT5 on the language-specific training data. The results are shown in mFT5* rows in Table \ref{tab:fmt5_coarse}. While zero-shot cross-lingual experiments did not provide the best results, they provided very competitive results compared to the baselines. The results confirm the strong cross-lingual nature of the pre-trained mFT5 model in detecting offensive language. It should also be noted that an MLM approach similar to fBERT and HateBERT  needs labelled data to fine-tune and will not be able to provide zero-shot offensive language identification. Therefore, mFT5 is useful for low-resource languages where the training data is scarce. 

\subsection{Token-level Offensive Language Identification}

We also experimented with token-level offensive language identification in the datasets where the token-level labels are available; Korean and Sinhala. The input to the model is the text, and the output is the text with "[OFF]" placeholders in front of the offensive tokens. To evaluate the models, we used the macro F1 score computed on predicted offensive tokens and gold offensive tokens. We compared the results with token classification architecture in mBERT and XLM-R large models.

\begin{table}[!ht]
\centering
\scalebox{.9}{
\begin{tabular}{llc}
\hline
\bf Dataset & \textbf{Model} & \textbf{Macro F1} \\ \hline

 \multirow{ 4}{*}{Korean} &  \bf mFT5 & \bf 0.489\textpm0.004 \\ 
&  XLM-R & 0.466\textpm0.009 \\ 
& mBERT & 0.453\textpm0.013 \\
& mT5 & 0.311\textpm0.016 \\
\hline
 \multirow{ 4}{*}{Sinhala} &  \bf mFT5 & \bf 0.743\textpm0.009 \\ 
&  XLM-R & 0.723\textpm0.013 \\ 
& mT5 & 0.316\textpm0.023 \\
& mBERT & 0.00 \\

\bottomrule
\end{tabular}
}
\caption{The test set macro F$_1$ scores for token-level offensive language detection. Results are ordered by performance. The best results are shown in bold font.}
\label{tab:mft5_finegrained}
\end{table}

\noindent As shown in Table \ref{tab:mft5_finegrained}, mFT5 outperforms mBERT and XLM-R in all of the token-level offensive language identification datasets too. It should be noted that pre-trained models with task-specific heads such as toxic-bert will not be able to perform token-level tasks. However, our text-to-text approach in mFT5 provided state-of-the-art results at token-level too.

\section{Conclusion and Future Work}

Neural transformer models have outperformed previous state-of-the-art deep learning models across different NLP tasks including offensive language identification. Following impressive results in international benchmark competitions, domain-specific pre-trained neural transformers such as ToxicBERT, fBERT and HateBERT have been proposed for offensive language identification. As discussed in this paper, these models have limitations which makes it difficult extend them in to different datasets. We address these limitations by proposing FT5, a \textit{t5-large} model that has been trained using over $2$ million instances from the SOLID and CCTK datasets. The FT5 model achieves better results in both sentence-level and token-level tasks across different offensive language identification benchmarks.

This paper also introduced mFT5. To the best of our knowledge, mFT5 is the first pre-trained multilingual offensive language detection model. The model uses the \textit{mt5-large} and is trained using the same data used to train FT5. We show that the proposed FmT5 model achieves better results in both sentence-level and token-level tasks compared to the mBERT, XLM-R, and the vanilla mT5 model across different offensive language identification benchmark datasets. We show that the model performs consistently well across different languages and platforms. Furthermore, the model showed strong zero-shot cross-lingual results opening exciting new avenues for multilingual offensive language detection. 

In future work, we would like to extend the proposed FT5 model to the identification of offensive spans along with their targets using the recently released TBO dataset \cite{zampieri-etal-2023-target}. We believe that modelling targets and offensive expressions jointly is an important step towards improving explainability in offensive language identification systems. Another important direction we have been exploring is the computational efficiency. We have recently experimented with teacher-student architectures using knowledge distillation (KD) and we have shown that the use of KD results in lightweight models that are more computationally efficient and perform on par with larger models \cite{ranasinghe-zampieri-2023-teacher}. We would like to investigate teacher-student architectures using the proposed FT5 model. Finally, we are interested in the application of multilingual models to low-resource scenarios. Thousands of languages and dialects are spoken in the world, but research on offensive language identification is still (mostly) restricted to English and a few other high-resource languages. We believe that the release of mT5 will encourage research on offensive language identification models for low-resource languages, dialects, and other challenging linguistic scenarios such as code-mixed texts. 

\section*{Limitations}
Training a T5 model requires a large amount of computing resources. We noted that training a T5 model can take more GPU resources than training a BERT model such as fBERT \cite{sarkar2021fbert}. Therefore, we did not experiment with large T5 models such as T5-XL and T5-XXL. While these models might perform better than T5-Large models we experimented with, they would consume more resources, limiting their potential use cases. 

\section*{Ethics Statement}

FT5 and mFT5 are essentially T5 models for offensive language identification, which is trained on multiple publicly available datasets. We used multiple datasets referenced in this paper which were previously collected and annotated to evaluate the models. No new data collection has been carried out as part of this work. We have not collected or processed writers'/users' information, nor have we carried out any form of user profiling to protect users' privacy and identity. 

We believe that content moderation should be a trustworthy and transparent process applied to clearly harmful content so it does not hinder individual freedom of expression rights. We encourage research in automatically detecting offensive content on the web trough a trustworthy and transparent process. Using our proposed models for this purpose will alleviate the psychological burden for social media moderators who are exposed to large amounts offensive content while ensuring a more transparent moderation process. 


\section*{Acknowledgments}

We would like to thank the anonymous AACL-IJCNLP reviewers for their insightful feedback. We further thank the creators of the datasets used in this paper for making the datasets publicly available for our research. 

The computational experiments in this paper were conducted on an Aston EPS Machine Learning Server, funded by the EPSRC Core Equipment Fund, Grant EP/V036106/1.

\bibliography{custom}
\bibliographystyle{acl_natbib}




\end{document}